\documentclass[10pt,twocolumn,letterpaper]{article}

\usepackage{cvpr}              
\usepackage{graphicx}
\usepackage{amsmath}
\usepackage{amssymb}
\usepackage{booktabs}

\usepackage[pagebackref,breaklinks,colorlinks]{hyperref}

\usepackage[capitalize]{cleveref}
\crefname{section}{Sec.}{Secs.}
\Crefname{section}{Section}{Sections}
\Crefname{table}{Table}{Tables}
\crefname{table}{Tab.}{Tabs.}

\makeatletter
\@namedef{ver@everyshi.sty}{}
\makeatother
\usepackage{tikz}
\usepackage{etoolbox}
\usepackage{enumitem}
\newcommand*{\circled}[1]{\lower.7ex\hbox{\tikz\draw (0pt, 0pt)%
    circle (.5em) node {\makebox[1em][c]{\small #1}};}}
\robustify{\circled}

\def\cvprPaperID{2513} 
\def\confName{CVPR}
\def\confYear{2022}

\begin{document}

\title{Depth-Aware Generative Adversarial Network \\ for Talking Head Video Generation}

\author{Fa-Ting Hong$^1$ \and Longhao Zhang$^2$ \and Li Shen$^2$ \and Dan Xu$^1$\thanks{Corresponding author} 
\vspace{3pt}
\and
$^1$Department of Computer Science and Engineering, HKUST\quad 
$^2$Alibaba Cloud\\
{\tt\small fhongac@cse.ust.hk, longhao.zlh@alibaba-inc.com, lshen.lsh@gmail.com, danxu@cse.ust.hk}
}
\maketitle

\begin{abstract}
Talking head video generation aims to produce a synthetic human face video that contains the identity and pose information respectively from a given source image and a driving video.~Existing works for this task heavily rely on 2D representations (e.g.~appearance and motion) learned from the input images. However, dense 3D facial geometry (e.g.~pixel-wise depth) is extremely important for this task as it is particularly beneficial for us to essentially generate accurate 3D face structures and distinguish noisy information from the possibly cluttered background. Nevertheless, dense 3D geometry annotations are prohibitively costly for videos and are typically not available for this video generation task. In this paper, we first introduce a self-supervised geometry learning method to automatically recover the dense 3D geometry (i.e.~depth) from the face videos without the requirement of any expensive 3D annotation data. Based on the learned dense depth maps, we further propose to leverage them to estimate sparse facial keypoints that capture the critical movement of the human head. In a more dense way, the depth is also utilized to learn 3D-aware cross-modal (i.e.~appearance and depth) attention to guide the generation of motion fields for warping source image representations. All these contributions compose a novel depth-aware generative adversarial network (DaGAN) for talking head generation. Extensive experiments conducted demonstrate that our proposed method can generate highly realistic faces, and achieve significant results on the unseen human faces.~\footnote{\url{https://github.com/harlanhong/CVPR2022-DaGAN}}


   
\end{abstract}

\vspace{-0.1cm}
\section{Introduction}
\vspace{-0.1cm}
In this paper, we target the task of generating a realistic talking head video of a person using a source image of that person and a driving video, possibly derived from another person~\cite{wu2018reenactgan, xu2017face,yao2020mesh}. In the real world, a wide range of practical applications can be benefited from this task such as role-playing video games and virtual anchors. 
\begin{figure}[t]
  \centering
    \includegraphics[width=0.9\columnwidth]{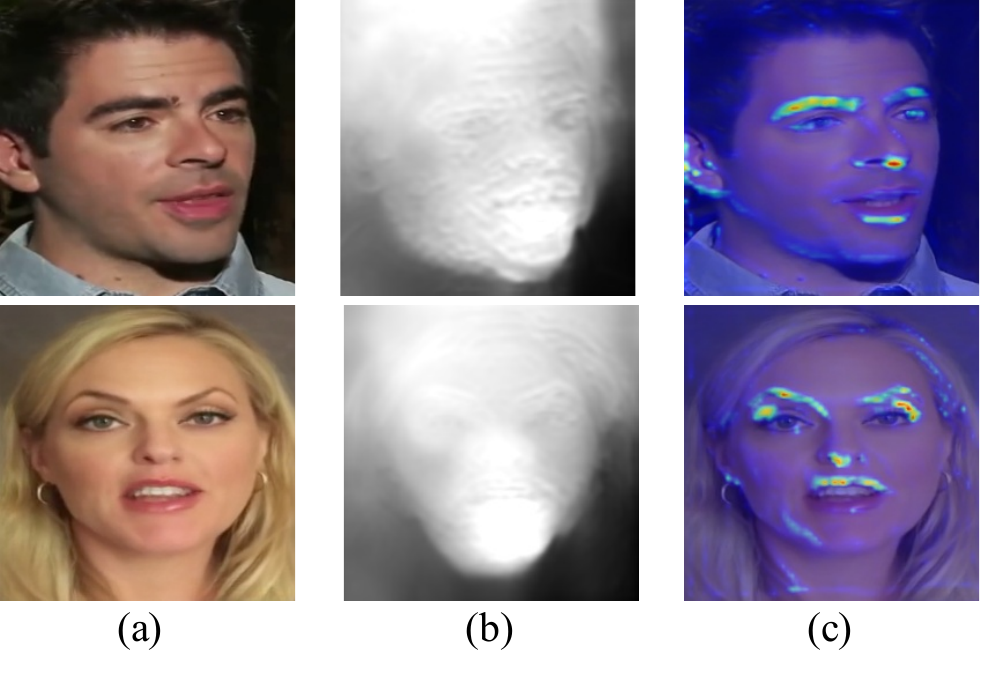}
    \vspace{-14pt}
\caption{Qualitative results of the learned depth maps (Fig.~\ref{fig:firstpage}b) of the face images (Fig.~\ref{fig:firstpage}a) using a self-supervised manner, and dense depth-aware attention maps (Fig.~\ref{fig:firstpage}c), which can attend to important semantic parts of the face such as eyes.
    }
    \vspace{-15pt}
    \label{fig:firstpage}       
\end{figure}
\par Rapid progress has been achieved on talking head video generation in terms of both quality and robustness in recent years, using generative adversarial networks (GANs)~\cite{goodfellow2014generative}.
A successful direction for the task in the literature focuses on decoupling identity and pose information from the face images~\cite{siarohin2019first, wang2021one,yao2020mesh}. 
For instance, pioneering works~\cite{siarohin2019first, wang2021one} propose to model relative poses between two face images based on estimated sparse facial keypoints, and the poses are further used to generate dense motion fields, which warps the feature maps of the source image to drive the image generation. Similarly, 
Eurkov~\etal~\cite{burkov2020neural} aimed to specifically learn two latent codes for the pose and the identity, and then input them into a designed generator network for face video synthesis. More than that, data augmentation strategies~\cite{burkov2020neural, zhou2021pose} are also explored to more effectively perform the disentanglement of the pose and identity information. Although these methods show highly promising performance on the task, they still pay large attention to learning more representative 2D appearance and motion features from the input images. However, for face video generation, 3D dense geometry is critically important for the task while rarely investigated in the existing methods.

\par The dense 3D geometry (\eg~pixel-level depth) can bring several significant benefits for the talking-head video generation. First, as the video captures the moving heads in a realistic 3D physical world, the 3D geometry can greatly facilitate an accurate recovery of 3D face structures, and the model capability of maintaining a realistic 3D face structure is a key factor for generating high-fidelity face videos. Second, the dense geometry can also help the model to robustly distinguish the noisy background information for generation especially under cluttered background conditions. Finally, the dense geometry is also particularly useful for the model to identify expression-related micro-movements on the faces. However, a severe issue of utilizing the 3D dense geometry to significantly boost the generation is that the 3D geometry annotations are highly expensive and typically not available for this task. 

To address this problem, in this paper, we first propose to learn the pixel-wise depth map (see Fig.~\ref{fig:firstpage}b) via geometric warping and photometric consistency in a \emph{self-supervised} manner, to automatically recover \emph{dense} 3D facial geometry from the training face videos, without requiring any expensive 3D geometry annotations. Based on the learned dense facial depth maps, we further propose two mechanisms to effectively leverage the depth information for better talking-head video generation. The first mechanism is depth-guided facial keypoint detection. The facial keypoints estimated by the network should well reflect the structure of the face, as they are further used to produce the motion field for feature warping, while the depth map explicitly indicates the 3D structure of the face. Thus, we combine geometry representations learned from the input depth maps with the appearance representations learned from the input images, to predict more accurate facial keypoints. The second mechanism is a cross-modal attention mechanism to guide the learning of the motion field. The motion field may contain noisy information from the cluttered background, and cannot effectively capture the expression-related micro-movements as they are generated from sparse facial keypoints. Therefore, we propose to learn depth-aware attention to have pixel-wise 3D geometry constraint on the motion field (see Fig.~\ref{fig:firstpage}c), to drive the generation with more fine-grained details of facial structure and movements.  
\par 
All the above-illustrated contributions compose a \textbf{D}epth-\textbf{a}ware \textbf{G}enerative \textbf{A}dversarial \textbf{N}etwork (\textbf{DaGAN}) to advance talking head video generation. 
Extensive experiments are conducted to qualitatively and quantitatively evaluate the proposed DaGAN model on two different datasets, \ie~VoxCeleb1~\cite{nagrani2017voxceleb} and CelebV~\cite{wu2018reenactgan}. 
The experimental results show that our proposed self-supervised depth learning strategy can produce accurate depth maps on both the source and the target human face images.
Our DaGAN model can also generate higher-quality face images compared with state-of-the-art methods. More specifically, our model is able to better preserve facial details, yielding a synthesized face with a more accurate expression and pose. 

\par In summary, the main contribution is three-fold:
\vspace{-3pt}
\begin{itemize}
\setlength{\itemsep}{0pt}
\setlength{\parsep}{0pt}
\setlength{\parskip}{3pt}
    \item To the best of our knowledge, we are the first to introduce a self-supervised learning method to recover explicit dense 3D geometry (\ie~depth maps) from face videos for talking head video generation, and utilize the learned depth to boost the performance. 
\item We propose a novel depth-aware generative adversarial network for talking head generation, which effectively incorporates the depth information into the generation network via two carefully designed mechanisms, \ie~depth-guided facial keypoint estimation, and cross-modal (\ie~depth and image) attention learning.

    \item Extensive experimental results show accurate depth recovery of face images and also achieve superior generation performance compared with state-of-the-arts.
    \vspace{-0.3cm}
\end{itemize}

\label{sec:intro}
\vspace{-0.1cm}
\section{Related Works}
\vspace{-0.1cm}
\noindent\textbf{Generative Adversarial Networks.} 
The generative adversarial network (GAN) was introduced by Goodfellow~\etal~\cite{goodfellow2014generative} to produce realistic images under certain conditions. GANs have attracted substantial attention and  has been studied in many tasks~\cite{liu2021generative}, such as image synthesis~\cite{radford2015unsupervised, karras2019style, karras2020analyzing,goodfellow2014generative}, text-to-image translation~\cite{reed2016generative, zhang2017stackgan}, and image inpainting~\cite{liu2021pd, jam2021r,li2019boosted}. 
In this work, we focus on
talking head video generation with GAN guided by 3D facial depth maps learned without any ground-truth depths.

\begin{figure*}[t]
  \centering
    \includegraphics[width=0.95\textwidth]{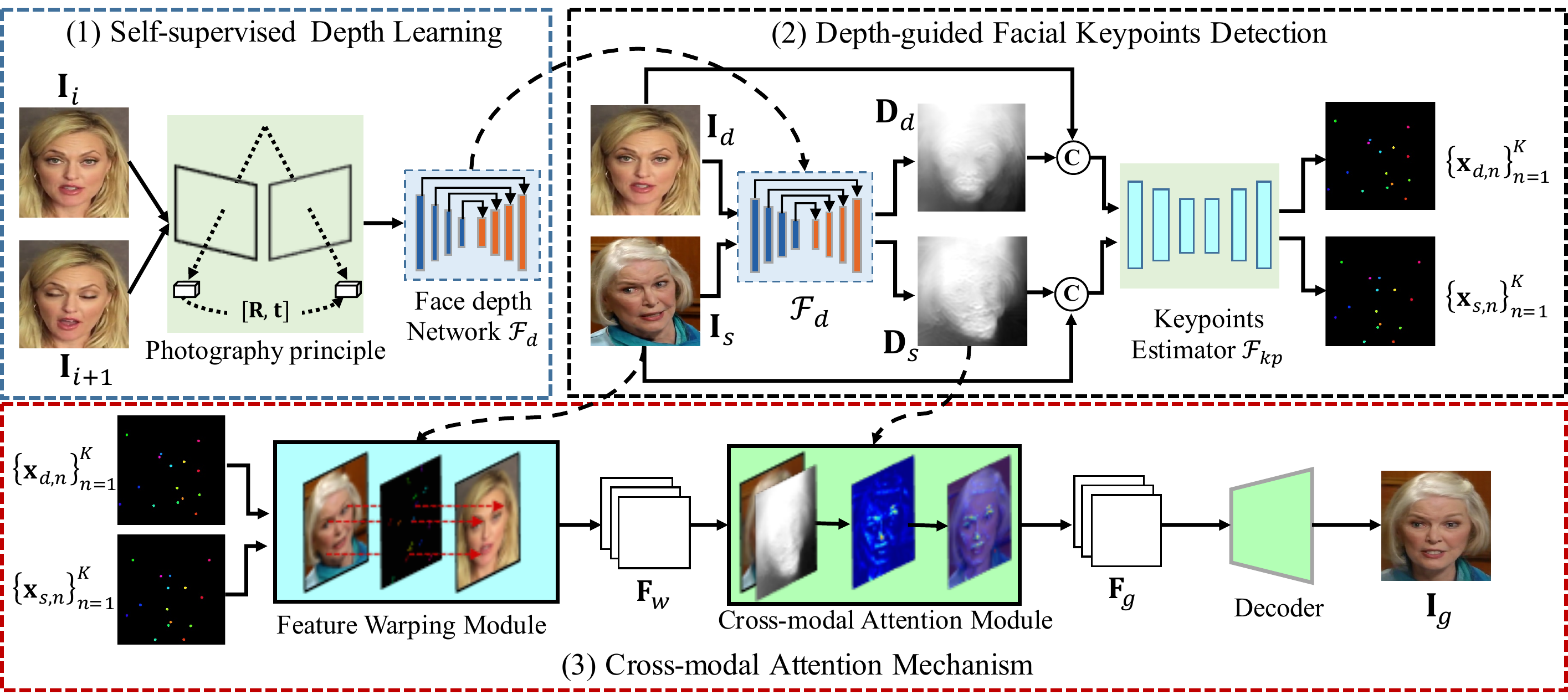}
    \vspace{-10pt}
    \caption{An illustration of the proposed DaGAN approach, which can be mainly divided into three sub-networks: (1) a self-supervised depth learning sub-network $\mathcal{F}_d$. We learn pixel-wise face depth maps in a self-supervised manner to recover the dense 3D facial geometry from the training face videos.
    (2) a depth-guided facial keypoints detection sub-network $\mathcal{F}_{kp}$. In this part, we combine both the geometry representations from depth maps with the appearance representations from the images to predict more accurate facial keypoints.
    (3) a cross-modal (\ie~depth and rgb image) attention learning sub-network. We learn dense depth-aware attention map using depth maps to constrain the motion field, to obtain a more accurate generation of fine-grained details of facial structure and movements.  
    }
    \vspace{-10pt}
    \label{fig:framework}       
\end{figure*}

\noindent\textbf{Depth Estimation.} 
Many works have been proposed to tackle the problem of depth estimation from stereo images or video sequences~\cite{luo2020consistent, digging2019monodepth2,fu2018deep,zhou2017unsupervised,ha2016high}.
Zhou~\etal~\cite{zhou2017unsupervised} use an end-to-end learning approach with view synthesis as the supervisory signal to estimate the depth map in monocular video sequences in an unsupervised manner. Based on \cite{zhou2017unsupervised}, Clement~\etal~\cite{digging2019monodepth2} gain a significant improvement using a minimum reprojection loss to deal with occlusions between frames and an auto-masking loss to ignore confusing stationary pixels. Gordon~\etal~\cite{gordon2019depth} tried to learn camera intrinsics for every two consecutive frames to make the model able to perform inference in the wild.

\par {However, our work aims to learn facial depth maps in an unsupervised manner for the talking head generation task with only video images required.} The depth map can provide dense 3D geometric information for the keypoint detection and can serve as an important cue to guide the model to focus on fine-grained critical parts of the human face (\eg~eyes, and mouth) during image generation.

\noindent\textbf{Talking Head Video Generation.}
Talking head video generation can be divided into three major strategies according to its driven-modality, \ie~image-driving methods~\cite{wang2021one, siarohin2019first, wiles2018x2face, zhang2019one, burkov2020neural, yao2020mesh}, landmark-driving methods~\cite{zakharov2019few, zakharov2020fast, zhao2021sparse, ha2020marionette} and audio-driving methods~\cite{zhou2019talking, wang2021audio2head, deng2020disentangled, zhou2021pose}. To exclude the driving face's identity information, several image-driving methods~\cite{siarohin2019first, wang2021one} tried to predict keypoints of both the source image and driving image, and model local motion using changes in the positions of corresponding keypoints.
Using facial landmarks instead of pure images to encode the pose information is an intuitive method. The fs-vid2vid~\cite{zakharov2019few} models person appearance by decomposing it into two layers, \ie~a pose-dependent coarse image and a pose-independent texture image. Zhao~\etal~\cite{zhao2021sparse} not only model global motions using full facial landmarks, but also use local landmarks to enforce the model to focus on local regions. The audio-driving method is a more popular way to perform face reenactment since the audio does not contain identity information, which can enable the model to more easily obtain a latent code of pose information from the audio. In~\cite{zhou2019talking}, the encoder disentangles the pose information from identity information assisted by the audio modality. 

In contrast to these existing works, we learn explicit pixel-wise depth map in a self-supervised manner, to provide highly beneficial 3D dense geometry information of the human faces, which allows the proposed model to accurately perceive 3D structures of the faces, and generate more fine-grained details of face spatial structures.
\vspace{-0.1cm}
\section{The proposed DaGAN Approach}
\vspace{-0.1cm}
Generating talking head videos is a technically challenging task as it requires the preserving of the identity information while imitating the facial motion from the driving faces. In this work, under the same setting as utilized in previous works~\cite{siarohin2019first, yao2020mesh}, we propose a depth-aware generative adversarial network, termed as DaGAN, for talking head video generation. It learns a depth estimation network in a self-supervised manner from training face videos, without requiring any expensive 3D geometry data as input. Thus, we can recover reliable face depth maps for both the input source and driving images to capture accurate 3D face structures and the expression-related micro-movements for higher-quality talking-head video generation. 
\vspace{-0.1cm}
\subsection{Overview} \label{sec:overview}
\vspace{-0.1cm}
 Our proposed DaGAN approach consists of a generator and a discriminator. The core network architecture of our generator is depicted in Fig.~\ref{fig:framework}, while the implementation of the discriminator is directly inspired from FOMM~\cite{siarohin2019first}. 
Our generation network can be split into three parts: (i) a self-supervised depth learning sub-network $\mathcal{F}_d$. 
The face depth network $\mathcal{F}_d$ first learns depth estimation using two consecutive frames (\ie~$\mathbf{I}_i$ and $\mathbf{I}_{i+1}$) from a face video in a self-supervised manner. Then the whole deep framework is jointly trained while with $\mathcal{F}_d$ fixed.~(ii) a depth-guided sparse keypoints detection sub-network $\mathcal{F}_{kp}$. Given a source image $\mathbf{I}_s$ and a driving image $\mathbf{I}_d$ from the driving video, we exploit $\mathcal{F}_d$ to produce depth maps ($\mathbf{D}_{s}$ and $\mathbf{D}_{d}$) for each image. These depth maps and their RGB images are concatenated to learn geometry and appearance features for detecting face keypoints (\ie $\{\mathbf{x}_{s,n}\}_{n=1}^N$ and $\{\mathbf{x}_{d,n}\}_{n=1}^N$), which can be used to generate relative motion fields of the human faces; 
(iii) the feature warping module accepts the keypoints as input to generate motion fields, which are used to warp the source-image feature map to fuse the motion with the appearance information, resulting in a warped feature $\mathbf{F}_w$. 
To enforce the model to focus on fine-grained details of face structures and micro-expression movements, we further learn a dense depth-aware attention map using the source depth map $\mathbf{D}_s$ and the warped feature $\mathbf{F}_w$. The depth-aware attention map can be used to refine the warped feature to produce a refined feature $\mathbf{F}_{g}$, resulting in a better generated image $\mathbf{I}_g$.

\begin{figure}[t]
  \centering
    \includegraphics[width=0.95\columnwidth]{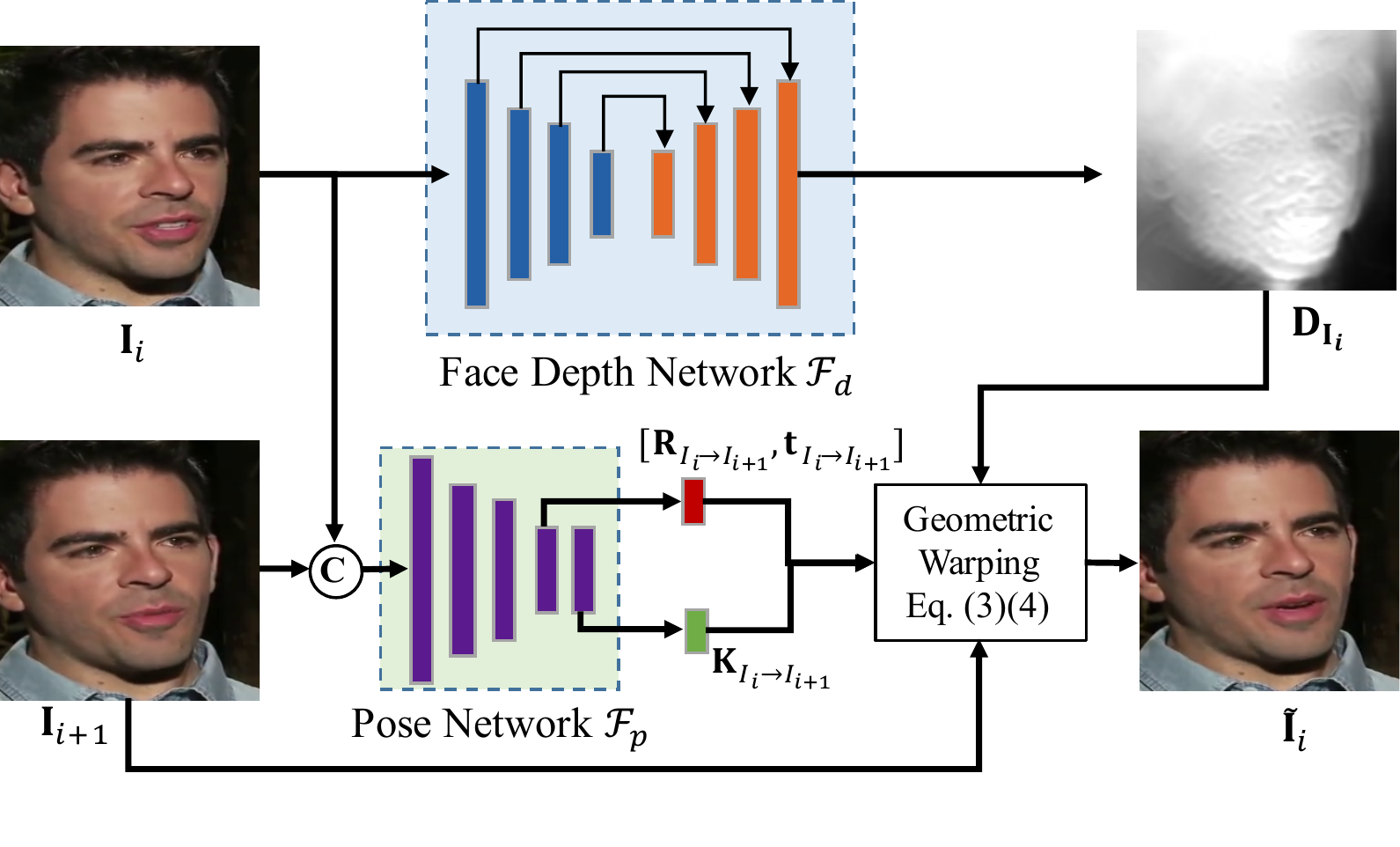}
    \vspace{-15pt}
    \caption{The training process of our face depth network. In addition to the face depth network, we use a pose network to estimate the relative camera poses [$\mathbf{R}_{I_i\rightarrow I_{i+1}},  \mathbf{t}_{I_i\rightarrow I_{i+1}}]$ and the camera intrinsic matrix $\mathbf{K}_{I_i\rightarrow I_{i+1}}$. The symbol {\footnotesize $\circled{c}$} represents the concatenated operation.}
    \label{fig:depth_estimate}  
    \vspace{-15pt}
\end{figure}
\vspace{-0.1cm}
\subsection{Self-supervised Face Depth Learning} \label{sec:depth learning}
\vspace{-0.1cm}
In this part, we elaborate the technical details of the proposed self-supervised facial depth learning network, which can automatically recover dense face depth maps from the input source and driving images. 
Although SfM-Learner~\cite{zhou2017unsupervised} previously proposed to learn \emph{outdoor} scene depth in an unsupervised manner in an autonomous driving scenario, while in this work, we extend the method to learn face depths specifically for talking head video generation. Since the facial videos contain relatively larger-area dynamic motion (moving head dominating on the image) compared to the outdoor scenes, unsupervised facial depth estimation is a challenging problem in our task.  

\par We optimize the depth network using available training face videos. Specifically, given two consecutive video frames $\mathbf{I}_i$ and $\mathbf{I}_{i+1}$ from a face video, with $\mathbf{I}_{i+1}$ as a source image and $\mathbf{I}_{i}$ as a target image, we aim to learn several geometric elements, including a depth map $\mathbf{D}_{I_i}$ for the target image $\mathbf{I}_{i}$, a camera intrinsic matrix $\mathbf{K}_{I_i\rightarrow I_{i+1}}$, and a relative camera pose $\mathbf{R}_{I_i\rightarrow I_{i+1}}$ with translation $\mathbf{t}_{I_i\rightarrow I_{i+1}}$ between the two images. It should be noted that the camera intrinsic $\mathbf{K}_{I_i\rightarrow I_{i+1}}$ is also not available in our training face video dataset, which is clearly different from~\cite{zhou2017unsupervised} directly using provided camera intrinsic parameters for geometric warping. $\mathbf{K}_{I_i\rightarrow I_{i+1}}$ is input-video-clip specifically learned in our method, as each face video can be possibly captured by any camera. So the input of our method only requires video frames. 

\par The depth map $\mathbf{D}_{I_i}$ can be produced using the depth network $\mathcal{F}_d(\cdot)$. The pose $\mathbf{R}_{I_i\rightarrow I_{i+1}}$, the translation $\mathbf{t}_{I_i\rightarrow I_{i+1}}$, and the camera intrinsic matrix $\mathbf{K}_{I_i\rightarrow I_{i+1}}$ are predicted from the same pose network $\mathcal{F}_p(\cdot)$ as follows:
\begin{gather}
    \mathbf{D}_{I_i} =  \mathcal{F}_d(\mathbf{I}_i), \\
    [\mathbf{R}_{I_i\rightarrow I_{i+1}}, \mathbf{t}_{I_i\rightarrow I_{i+1}}],
    \mathbf{K}_{I_i\rightarrow I_{i+1}} = \mathcal{F}_p(\mathbf{I}_i \ || \ \mathbf{I}_{i+1}), 
\end{gather}
where the symbol $||$ indicates a concatenation of the two images.
Then, we can warp the source image $\mathbf{I}_{i+1}$ to the view of the target image $\mathbf{I}_{i}$ as follows:
\begin{gather}
    \mathbf{q}_k \sim
    \mathbf{K}_{I_i\rightarrow I_{i+1}} [\mathbf{R}_{I_i\rightarrow I_{i+1}} \, | \, \mathbf{t}_{I_i\rightarrow I_{i+1}}] \mathbf{D}_{I_i}(\mathbf{p}_j)\mathbf{K}_n^{-1}\mathbf{p}_j,  \\
    \mathbf{\widetilde{I}}_{i} = \mathcal{B_I}(\mathbf{I}_{i+1}, \{\mathbf{q}_{k}\}_{k=1}^N),
\end{gather}
where $\mathbf{q}_{k}$ and $\mathbf{p}_{j}$ denote the warped pixel on the source image $\mathbf{I}_{i+1}$ and an original pixel on the target image $\mathbf{I}_{i}$; $N$ is the overall number of pixels of the image; $\mathcal{B_I}(\cdot)$ is a differentiable bilinear interpolation function; $\mathbf{\widetilde{I}}_{i}$ is a reconstructed image at the source view. Therefore, we can construct a photometric consistency error $Pe(\cdot,\cdot)$ between $\mathbf{\widetilde{I}}_{i}$ and $\mathbf{I}_{i}$ to train our depth network in a self-supervised manner. Following~\cite{digging2019monodepth2}, we use L1 and SSIM~\cite{wang2004image} to construct the photometric consistency error $Pe$ as:
\begin{equation} \label{eq:pe_loss}
Pe(\mathbf{I}_{i}, \mathbf{\widetilde{I}}_{i}) = \alpha (1-SSIM(\mathbf{I}_{i}, \mathbf{\widetilde{I}}_{i})) 
+ (1-\alpha)||\mathbf{I}_{i} - \mathbf{\widetilde{I}}_{i}||,
\end{equation}
where $\alpha$ is set to 0.8 which shows better optimization in our experiments. After training the framework, we only utilize the face depth network $\mathcal{F}_d$ in DaGAN to estimate the depth maps of input face images, which are further employed by our proposed mechanisms for talking head generation.

\vspace{-0.1cm}
\subsection{Motion Modeling by Sparse Keypoints}\label{sec:feature warping}
\vspace{-0.1cm}
After we obtain the depth map from the face depth network, we concatenate the RGB image and its corresponding depth map produced by $\mathcal{F}_d$. Then, the keypoints estimator $\mathcal{F}_{kp}$ accepts the concatenated appearance (\ie~$\mathbf{I}_\tau$) and geometry (\ie~$\mathbf{D}_\tau$) information as inputs to more accurately predict a set of sparse keypoints of the human face:
\begin{equation}
    \{\mathbf{x}_{\tau,n}\}_{n=1}^K = \mathcal{F}_{kp}(\mathbf{I}_\tau \ || \ \ \mathbf{D}_\tau), \tau\in\{s, d\},
\end{equation}
where $K$ is the number of the detected face keypoint, and the subscript $\tau$ indicates a source image or a driving image; $||$ denotes a concatenation operation. We follow the design of~\cite{siarohin2019first} to implement our keypoints detector.

We adopt a feature warping strategy to capture head movements between the source and the target images, and implement a proposed feature warping module. 
Firstly, we compute a set of initial 2D offsets $\{\mathbf{O}_n\}_{n=1}^K$ for all the keypoints as follows:
\begin{equation}
    \{\mathbf{O}_n\}_{n=1}^K = \{\mathbf{x}_{s,n}\}_{n=1}^K - \{\mathbf{x}_{d,n}\}_{n=1}^K.
\end{equation}
Then, we generate a 2D dense coordinate map $z$ similar to~\cite{siarohin2019first}. After that, a dense 2D motion field $\mathbf{w}_m$ is generated by adding the $K$ offsets $\{\mathbf{O}_n\}_{n=1}^K$ into the 2D coordinate map at the corresponding coordinates of the $K$ keypoints.
 \par
\begin{figure}[t]
  \centering
    \includegraphics[width=0.95\columnwidth]{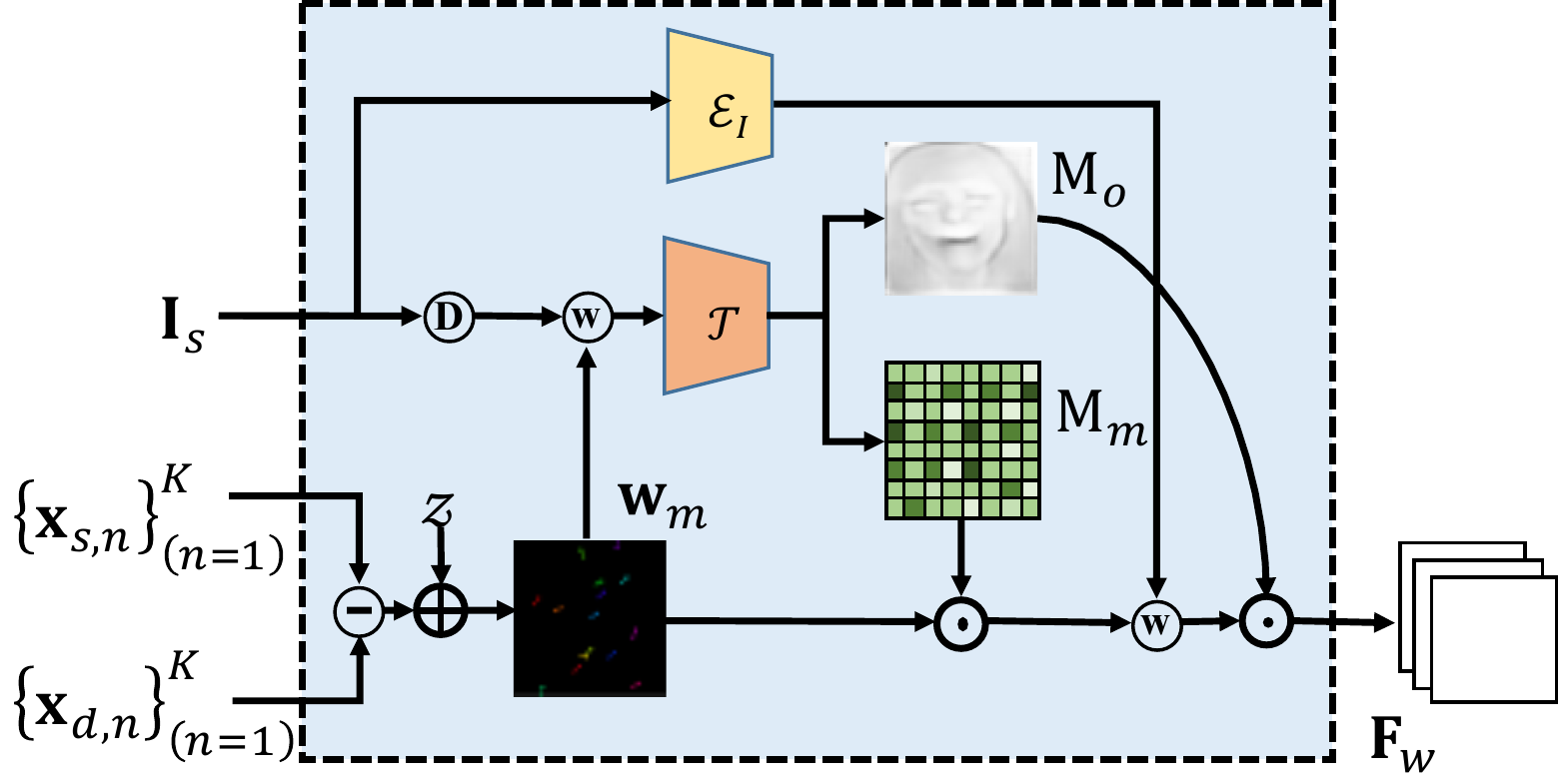}
    \vspace{-8pt}
    \caption{The illustration of our feature warping module. Here, {\footnotesize$\circled{\scriptsize{D}}$} is the downsampling operation, {\footnotesize$\circled{\scriptsize{w}}$} is the warping operation, {\footnotesize$\bigodot$} is the element-wise multiplication. The {\footnotesize$\bigoplus$} and {\footnotesize\circled{$\bf{-}$}} represent the addition and subtraction operation, respectively. 
    }
    \label{fig:feature_warp}   
    \vspace{-10pt}
\end{figure}

 As shown in Fig.~\ref{fig:feature_warp}, we first utilize the dense 2D motion field $\mathbf{w}_m$ to warp the downsampled image to produce an initial warped feature map. After that, an occlusion estimator $\mathcal{T}$ take as input the initial warped feature map to predict a motion flow mask $\mathbf{M}_m$ and an occlusion map $\mathbf{M}_o$~\cite{yao2020mesh}. The motion flow mask $\mathbf{M}_m$ assigns different confidence values for the estimated dense 2D motion field $\mathbf{w}_m$, resulting in masked motion field, while the occlusion map $\mathbf{M}_o$ aims to mask out the feature map regions that should be inpainted since the head has varying rotations. we utilize the masked motion field to warp the appearance feature map learned from the source image $\mathbf{I}_s$ extracted by the feature encoder $\mathcal{E}_I$. Then, they are fused with the occlusion map $\mathbf{M}_o$ to produce the warped soruce-image feature $\mathbf{F}_w$ as follows:
\begin{equation}
    \mathbf{F}_w = \mathbf{M}_o * \mathcal{W}_p(\mathcal{E_I}(\mathbf{I}_s), \mathbf{M}_m * \mathbf{w}_m),
\end{equation}
where $\mathcal{W}_p$ denotes the warping function. By so doing, the warped features $\mathbf{F}_w$ can better preserve the identity of the source image while maintaining the head motion information between two faces.

%
%

\vspace{-0.1cm}
\subsection{Cross-Modal Attention Module} \label{sec:Feature Refinement}
\vspace{-0.1cm}
To effectively embed the learned depth maps to boost the generation in a more dense way, we propose a cross-modal (\ie~depth and image) attention mechanism to enable the model to better preserve the facial structure and generate for expression-related micro facial movements, as the depth can provide us dense 3D geometry, which is essentially beneficial for maintaining the facial structure and identifying the critical movements when performing the generation. More specifically, we develop a cross-modal attention module to produce a dense depth-aware attention map to guide the warped feature $\mathbf{F}_w$ for face generation.   
\par
\begin{figure}[t]
  \centering
    \includegraphics[width=0.95\columnwidth]{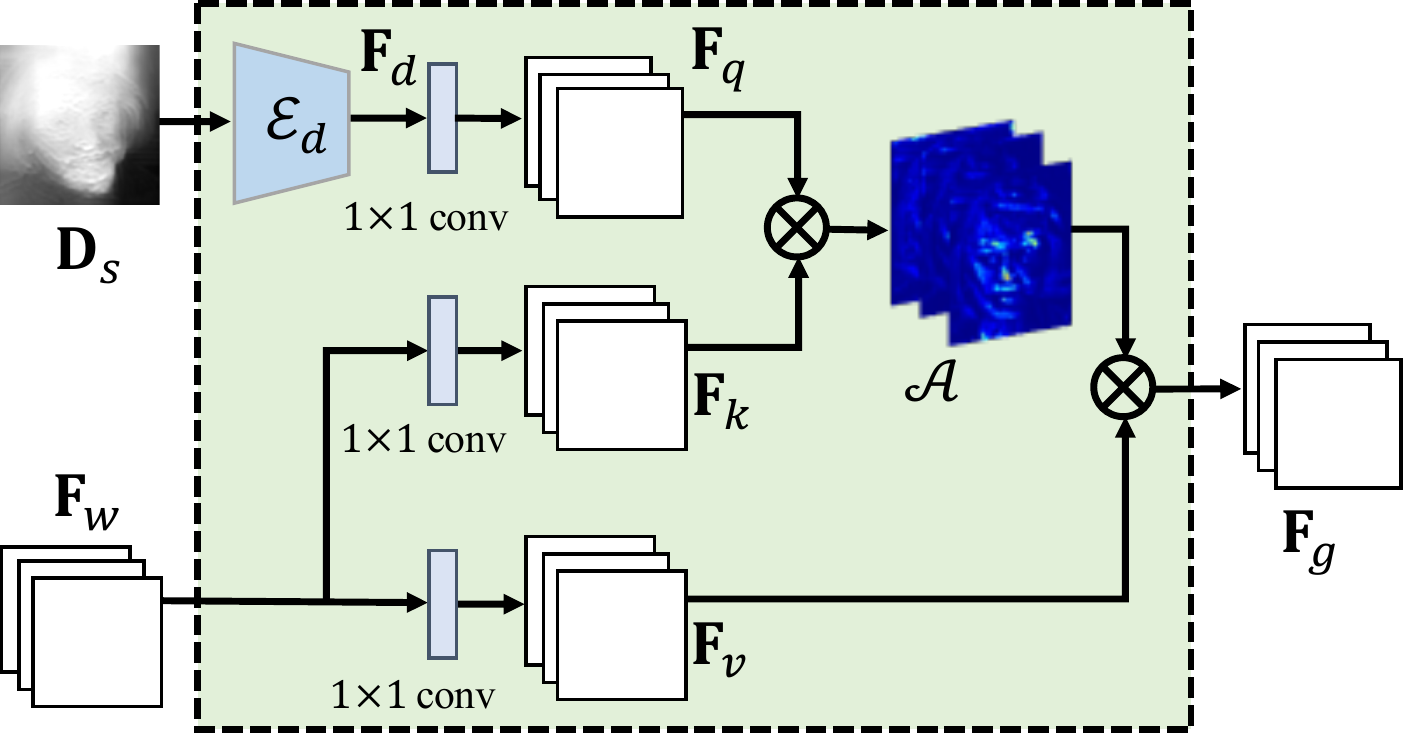}
    \vspace{-10pt}
    \caption{The illustration of our cross-modal attention module. Here, $1\times 1$ convolutional layers do not share the parameters with each other, and the symbol \footnotesize{$\bigotimes$} represents the matrix multiplication.}
    \label{fig:depth_attention}  
    \vspace{-10pt}
\end{figure}

 As shown in Fig. \ref{fig:depth_attention}, a depth encoder $\mathcal{E}_d$ take a source depth map $\mathbf{D}_s$ as input to encode a depth feature map $\mathbf{F}_d$, and we perform linear projection on $\mathbf{F}_d$ and the warped source-image feature $\mathbf{F}_w$ into three latent feature maps $\mathbf{F}_q$, $\mathbf{F}_k$ and $\mathbf{F}_v$ by three different $1\times 1$ convolutional layers with kernels $\mathbf{W}_q$, $\mathbf{W}_k$, and $\mathbf{W}_v$, respectively. The $\mathbf{F}_q$, $\mathbf{F}_k$ and $\mathbf{F}_v$ can respectively represent the query, key and value in the self-attention mechanism. Thus, the geometry-related query feature $\mathbf{F}_q$ produced by the depth map can be fused with the appearance-related key feature $\mathbf{F}_k$ to generate dense guidance for the human face generation. 
We obtain the final refined features  $\mathbf{F}_{g}$ for generation:
\begin{gather}
    \mathbf{F}_{g} =  \mathrm{Softmax}\left((\mathbf{W}_q\mathbf{F}_d) (\mathbf{W}_k\mathbf{F}_w)^T\right) \times (\mathbf{W}_v\mathbf{F}_w),
\end{gather}
where $\mathrm{Softmax}(\cdot)$ represents a softmax normalization function which outputs the dense depth-aware attention map $\mathcal{A}$ in Fig.~\ref{fig:depth_attention}. The $\mathcal{A}$ contains important 3D geometric guidance for generating the faces with more fine-grained details of facial structure and micro-movements. Finally, the decoder takes as input the refined warped features $\mathbf{F}_{g}$ to produce the final synthesized image $\mathbf{I}_g$.

%
%
\vspace{-0.1cm}
\subsection{Training}\label{sec:training}
\vspace{-0.1cm}
 In the training stage, the identities of the source and the driving image are the same, while they can be different in the inference stage. Following the previous works~\cite{siarohin2019first, wang2021one}, we train the proposed DaGAN in a self-supervised manner by minimizing the following loss:
 \begin{equation}
 \begin{aligned}
\mathcal{L} = &\lambda_P\mathcal{L}_P(\mathbf{I}_g,\mathbf{I}_d)+\lambda_G\mathcal{L}_G(\mathbf{I}_g,\mathbf{I}_d) \\
&+\lambda_E\mathcal{L}_E(\{\mathbf{x}_{d,n}\}_{n=1}^K)\\
&+ \lambda_D(\mathcal{L}_D(\{\mathbf{x}_{s,n}\}_{n=1}^K)+\mathcal{L}_D(\{\mathbf{x}_{d,n}\}_{n=1}^K)). 
 \end{aligned}
 \end{equation}
\noindent\textbf{Perceptual loss $\mathcal{L}_P$.} We minimize the perceptual loss~\cite{johnson2016perceptual} between the driving image $\mathbf{I}_d$ and the generated image $\mathbf{I}_g$, which has been effectively demonstrated being able to produce visually sharp outputs~\cite{siarohin2019first}. Moreover, we create an image pyramid for the driving image $\mathbf{I}_d$ and the generated image $\mathbf{I}_g$ to compute a pyramid perceptual loss.
\par\noindent\textbf{GAN loss $\mathcal{L}_G$.} We adopt the least-squares loss~\cite{mao2017least} as our adversarial loss. We use the discriminator to compute feature maps of different scales from the input image, and perform $\mathcal{L}_G$ on multiple levels as $\mathcal{L}_P$. We also minimize the discriminator feature matching loss~\cite{wang2021one}.

\noindent\textbf{Equivariance loss $\mathcal{L}_E$.} For a valid keypoint, when applying a 2D transformation to the image, the predicted keypoint should change according to the applied transformation~\cite{siarohin2019first}. Thus, we utilize an equivariance loss $\mathcal{L}_E$ to ensure the consistency of image-specific keypoints.

\noindent\textbf{Keypoints distance loss $\mathcal{L}_D$.} In order to make the detected facial keypoints aovid crowded around a small neighbourhood, we employ a keypoints distance loss to penalize the model if the distance of two corresponding keypoints falls below a predefined threshold.

Overall, the first two terms ensure the generated image being similar to the ground-truth. The third one enforces the predicted keypoints to be consistent, while the last one constrains the keypoints not to be clustered together. The $\lambda_P$, $\lambda_G$, $\lambda_E$ and $\lambda_D$ are the hyper-parameters to allow for a balanced learning from those losses. More details about the losses are presented in the {Supplementary Material}.

%
%
\vspace{-0.1cm}
\section{Experiments}
\vspace{-0.1cm}
In this section, we conduct extensive experiments on two talking face datasets to evaluate our proposed method. More additional experiments results and video samples are reported in the {Supplementary Material}.

%
%
\vspace{-0.1cm}
\subsection{Dataset and Metrics}
\vspace{-0.1cm}
\noindent\textbf{Dataset}. We mainly conduct experiments on two talking head generation datasets (\ie VoxCeleb1~\cite{nagrani2017voxceleb} dataset and CelebV~\cite{wu2018reenactgan} dataset) in this work. We follow the test set sampling strategy of MarioNETte~\cite{ha2020marionette}.

\noindent\textbf{Metrics}. In this work, several metrics are utilized to evaluate the quality of the generated images. Specifically, we use structured similarity (\textbf{SSIM}) and peak signal-to-noise ratio (\textbf{PSNR}) to evaluate the low-level similarity between the generated image and the driving image. Also, we adopt other three metrics, \ie \textbf{$\mathcal{L}_1$}, Average Keypoint Distance (\textbf{AKD}), and Average Euclidean Distance (\textbf{AED}) proposed in~\cite{siarohin2019animating} to evaluate the keypoint-based methods.

In cross-identity reenacting experiments, following the previous work~\cite{ha2020marionette}, we adopt the \textbf{CSIM} to evaluate the quality of identity preservation between source images and generated images. \textbf{PRMSE} is utilized to evaluate the head poses, while \textbf{AUCON} for expression evaluation. 

%
%
\vspace{-0.1cm}
\subsection{Implementation Details}
\vspace{-0.1cm}
The structure of the keypoints estimator is an hourglass network~\cite{yang2017stacked}. We use similar architectures as in~\cite{digging2019monodepth2} for implementing our depth and pose networks, while the decoder in the generator is the same as in~\cite{siarohin2019first}. The details of the structures of each sub-network in the proposed DaGAN is elaborated in {Supplementary Material}. For the optimization losses, we set $\lambda_P$ = 10, $\lambda_G$ =1, $\lambda_E$ = 10, and $\lambda_D$ = 10. We set the number of keypoints in DaGAN as $15$. In the training stage, we first train our face depth network using consecutive frames from videos in VoxCeleb1, and we fix it during the training of the whole deep generation framework. 

\vspace{-0.1cm}
\subsection{Comparison with State-of-the-art Methods}
\vspace{-0.1cm}
\begin{table}[t]
  \centering
  \resizebox{1\columnwidth}{!}{
        \begin{tabular}{ cccccc}
        \toprule
        Model  & CSIM $\uparrow$& SSIM$\uparrow$  & PSNR$\uparrow$ & PRMSE $\downarrow$ & AUCON$\uparrow$\\
        \midrule
        X2face~\cite{wiles2018x2face} & 0.689& 0.719 & 22.537 & 3.26 & 0.813\\
        NeuralHead-FF~\cite{zakharov2019few} & 0.229& 0.635 & 20.818 & 3.76 & 0.719\\
        MarioNETte~\cite{ha2020marionette} & 0.755 & 0.744 & 23.244 & 3.13 & 0.825\\
        FOMM~\cite{siarohin2019first}  & 0.813& 0.723 & 30.394 & 3.20 & 0.886\\
        MeshG~\cite{yao2020mesh} & 0.822 & 0.739 & 30.394 & 3.20 & 0.887 \\
        OSFV~\cite{wang2021one} & \underline{0.895} & \underline{0.761} & \underline{30.695} & \underline{1.64} & \underline{0.921} \\
        \midrule
        DaGAN (ours) & \textbf{0.899} & \textbf{0.804} & \textbf{31.220} & \textbf{1.22} & \textbf{0.939} \\
        \bottomrule
        \end{tabular}
}
\vspace{-8pt}
  \caption{Comparisons with state-of-the-art methods on the self-reenactment on the VoxCeleb1 dataset~\cite{nagrani2017voxceleb}. $\uparrow$ indicates larger is better, while $\downarrow$ indicates smaller is better.}
\label{tab:vox1}
\vspace{-6pt}
\end{table}

\begin{table}[t]
  \centering
  \resizebox{0.70\columnwidth}{!}{
        \begin{tabular}{cccc}
        \toprule
        Model & $\mathcal{L}_1$ $\downarrow$ & AKD $\downarrow$ & AED$\downarrow$\\
        \midrule
        X2face~\cite{wiles2018x2face}& 0.078& 7.687 & 0.405\\
        Monkey-Net~\cite{siarohin2019animating} & 0.049 & 1.878 & 0.199\\
        FOMM~\cite{siarohin2019first}  & \underline{0.043} & \underline{1.294} & \underline{0.140}\\
        OSFV~\cite{wang2021one}  & \underline{0.043} & 1.620 & 0.153\\
        \midrule
        DaGAN (ours) & \textbf{0.036} & \textbf{1.279}& \textbf{0.117} \\
        \bottomrule
        \end{tabular}
}
\vspace{-8pt}
  \caption{Comparisons with keypoint-based methods on self-reenactment on the VoxCeleb1 dataset~\cite{nagrani2017voxceleb}. $\downarrow$ smaller is better.}
\label{tab:vox1_akd}
\vspace{-6pt}
\end{table}

\begin{figure}[t]
  \centering
  \includegraphics[width=0.99\linewidth]{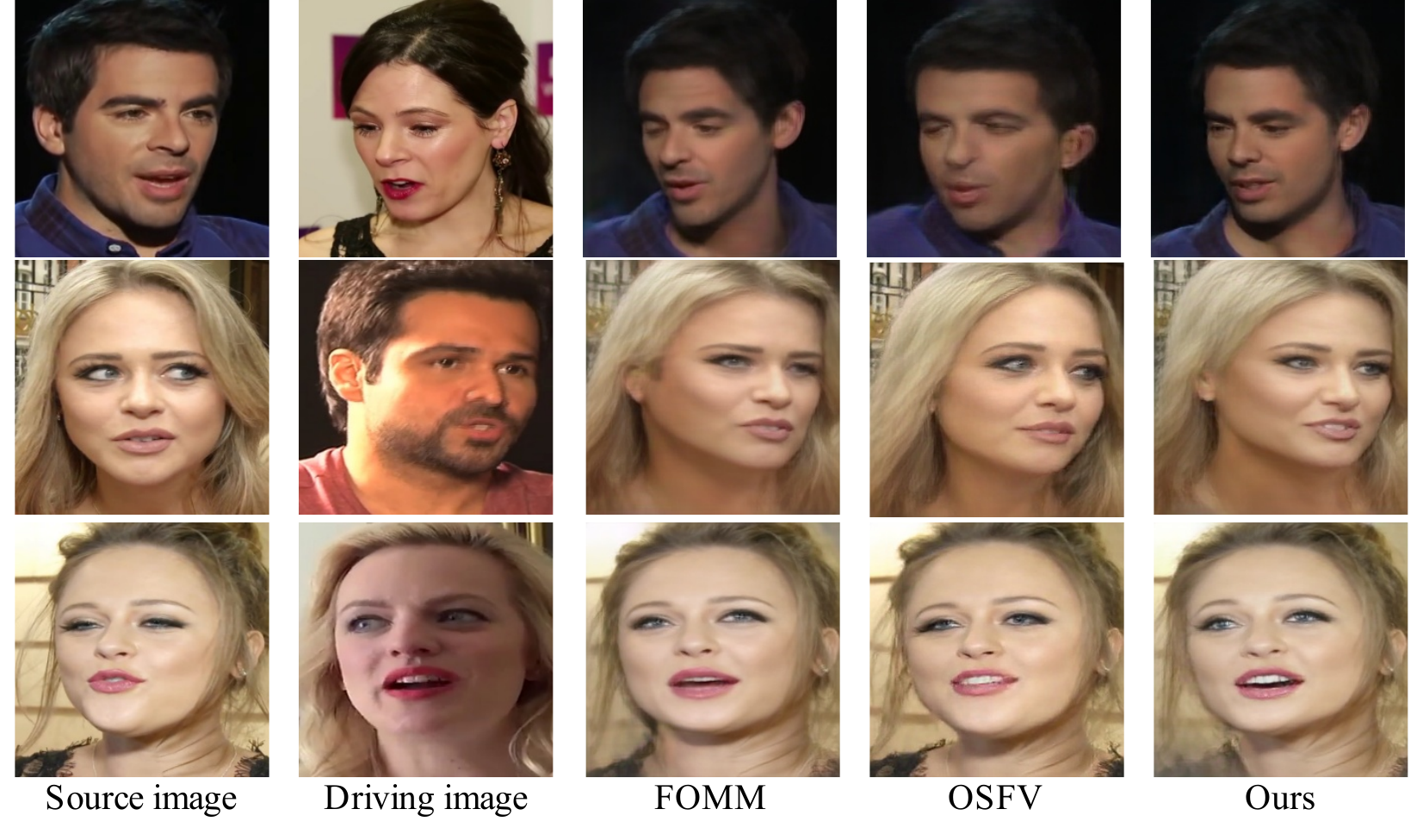}
  \vspace{-10pt}
   \caption{Qualitative comparisons of cross-identity reenactment on the VoxCeleb1 dataset~\cite{nagrani2017voxceleb}.}
   \label{fig:vox1}
   \vspace{-15pt}
\end{figure}
\noindent\textbf{Self-reenactment.} We first compare the face synthesis results where the source and driving images are of the same person, and report the results in Tab.~\ref{tab:vox1}. It can be observed that our DaGAN achieves the best results among all the compared methods. 
With a comparison to the other two keypoint-driven methods, \ie~FOMM~\cite{siarohin2019first} and OSFV~\cite{wang2021one}, our DaGAN model obtains the most accurate head movements ({1.22 of ours vs.~3.20 of FOMM}, resulting in \textbf{1.64} point improvement on the PRMSE metric), which verifies that our depth-guided facial-keypoints estimation can better capture the motion of human heads.~Regarding the facial expression, our method still obtains the highest score (\ie~$0.939$ on AUCON), meaning that our method can recover more fine-grained details of the face structures and micro-expression movements of the face. Also, our method produces the highest scores in both SSIM and PSNR, which demonstrates that our method can produce more realistic images compared with the most competitive methods. Additionally, we report the results on other three metrics proposed by~\cite{siarohin2019animating} in Tab.~\ref{tab:vox1_akd}. Our method obtains the best scores in these three metrics, clearly confirming our initial motivation that introducing the 3D depth maps can greatly benefit the keypoint-based generation. 

\begin{table}[t]
  \centering
  \resizebox{0.96\columnwidth}{!}{
        \begin{tabular}{ cccc}
        \toprule
        Model & CSIM$\uparrow$ & PRMSE $\downarrow$ & AUCON$\uparrow$\\
        \midrule
        X2face~\cite{wiles2018x2face}& 0.450& 3.62 & 0.679\\
        NeuralHead-FF~\cite{zakharov2019few} & 0.108 & 3.30 & 0.722\\
        marioNETte~\cite{ha2020marionette} & 0.520 & 3.41 & 0.710\\
        FOMM~\cite{siarohin2019first} & 0.462 & 3.90 & 0.667 \\
        MeshG~\cite{yao2020mesh} & 0.635 & 3.41 & 0.709 \\
        OSFV~\cite{wang2021one} & \textbf{0.791} & \underline{3.15} & \underline{0.805} \\
        \midrule
        DaGAN (ours) & \underline{0.723} & \textbf{2.33}& \textbf{0.873} \\
        \bottomrule
        \end{tabular}
}
\vspace{-0.25cm}
  \caption{Comparisons with state-of-the-art methods on cross-identity reenactment on CelebV dataset~\cite{wu2018reenactgan}. }
\label{tab:celebv}
\vspace{-0.3cm}
\end{table}

\begin{figure}[t]
  \centering
  \includegraphics[width=0.97\linewidth]{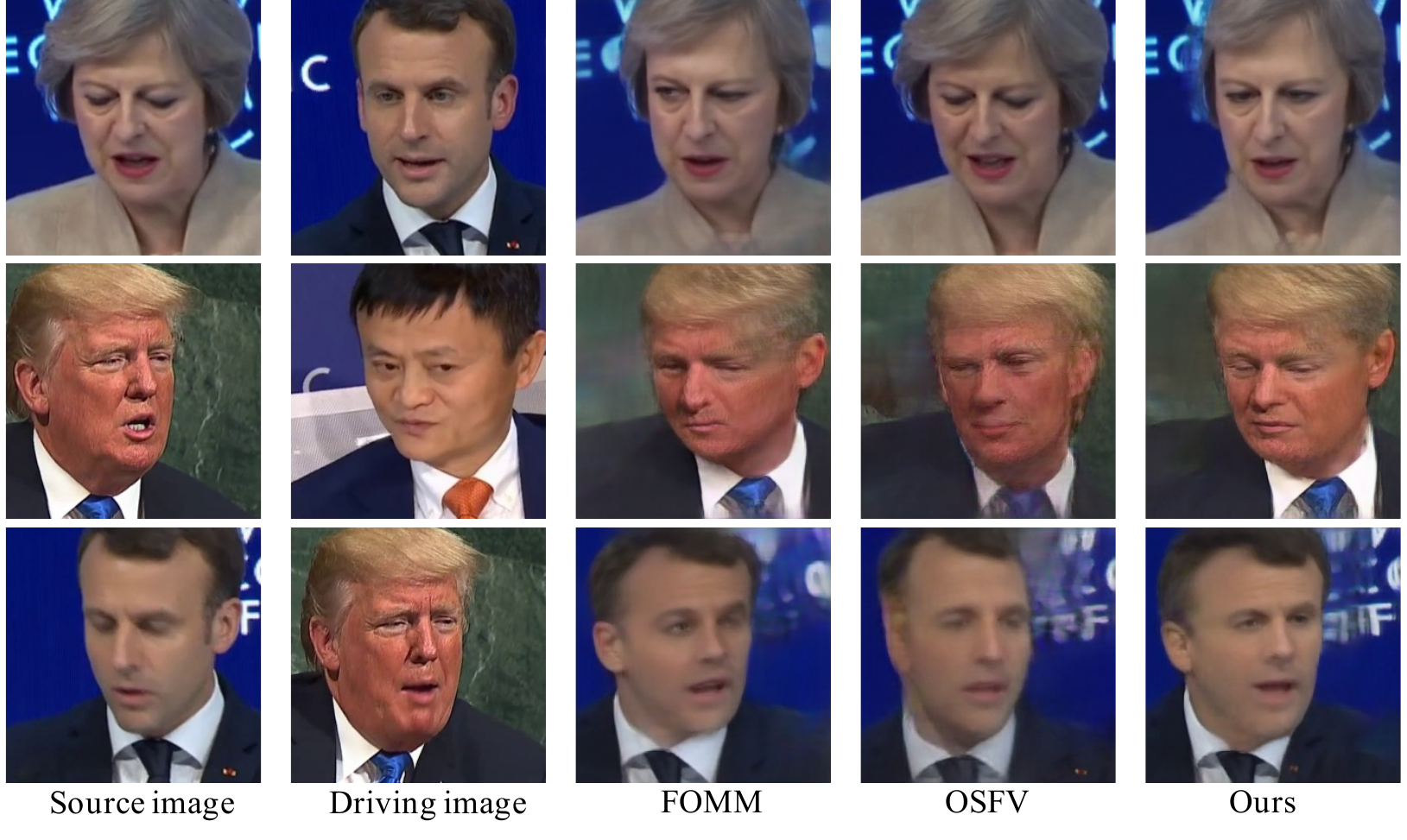}
  \vspace{-12pt}
   \caption{Qualitative comparisons of cross-identity reenactment on the CelebV dataset~\cite{wu2018reenactgan}.}
   \label{fig:celebv}
   \vspace{-16pt}
\end{figure}
\noindent\textbf{Cross-identity reenactment.} We also perform experiments on the CelebV dataset to exploit the cross-identity motion transfer, where the source and driving images are from different persons. We report the experimental results in Table~\ref{tab:celebv}. As we can observe that the PRMSE and AUCON of our DaGAN method remain the best among all methods, achieving 2.33 for PRMSE and 0.873 for AUCON. 
We also present several generated examples in Fig~\ref{fig:vox1} and~\ref{fig:celebv}. As some methods do not release their code, we only show the results of those methods with available codes (\eg~FOMM and OSFV). For the seen faces in Fig.~\ref{fig:vox1}, our method produces face images with more fine-grained details than the others. For instance, the mouth and eyes regions in three rows. It verifies that the utilization of depth maps enables the model to identify micro-expression movements of the human faces. For the unseen targets in the CelebV dataset, we also show some generated samples in~Fig.~\ref{fig:celebv}. Our method can also produce visually natural results for unseen targets. Notably, the generated images of OSFV in the first row is almost the same as the source image as it cannot detect the subtle motion on the face, which is also part of the reason why it outperforms our method in terms of CSIM in Tab.~\ref{tab:celebv}.

\begin{figure*}[ht]
  \centering
  \includegraphics[width=0.98\linewidth]{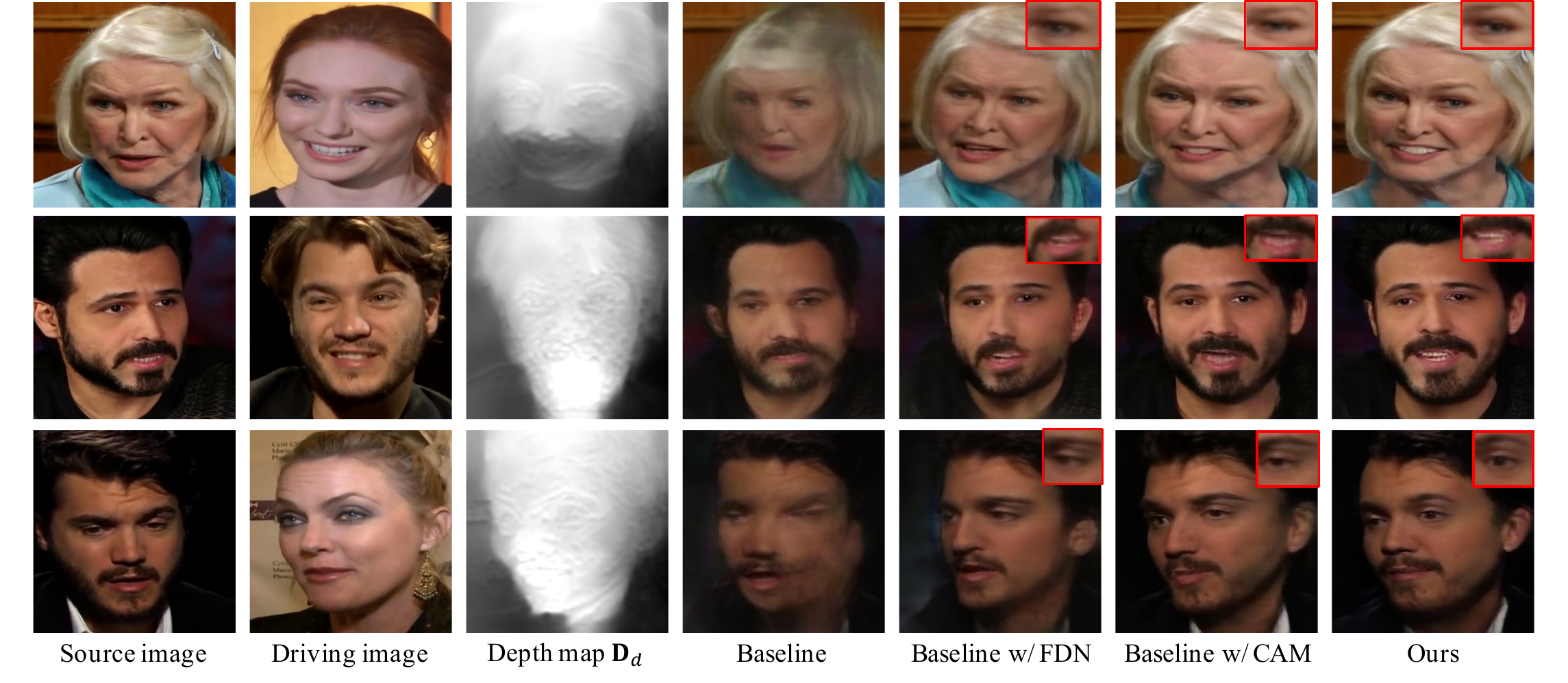}
  \vspace{-10pt}
   \centering\caption{Qualitative ablation studies. Depth map and depth attention module can obtain improvements compared with baseline, while our full method produce the most realistic image.}
   \vspace{-10pt}
   \label{fig:abla}
\end{figure*}
\begin{table}[t]
  \centering
  \resizebox{0.96\columnwidth}{!}{
        \begin{tabular}{ cccc}
        \toprule
        Model & CSIM$\uparrow$ & PRMSE $\downarrow$ & AUCON$\uparrow$\\
        \midrule
        Baseline & 0.688 & 5.39 & 0.657\\
        Baseline w/ FDN & 0.710 & 2.69 & 0.852\\
        Baseline w/ CAM & 0.698 & 2.56 & 0.838\\
        
        \midrule
        Ours (SA) & 0.681 & 5.18 & 0.832 \\
        DaGAN (ours) & \textbf{0.723} & \textbf{2.33}& \textbf{0.873} \\
        
        \midrule
        \midrule
        FOMM  & 0.462 & 3.90 & 0.667\\
        FOMM w/ FDN & 0.695 &  2.81 & 0.812\\
        FOMM w/ CAM & 0.669 & 2.36 & 0.821\\
        \midrule
        FOMM w/ FDN+CAM & \textbf{0.716} & \textbf{2.28} & \textbf{0.865}\\
        \bottomrule
        \end{tabular}
}
\vspace{-6pt}
  \caption{Ablation study. ``Baseline'' demotes the simplest model trained without the face depth network and cross-modal attention module. ``Baseline w/ CAM'' indicates that the baseline employs the cross-modal attention module after feature warping module, while ``Baseline w/ FDN'' combines the face depth network to estimate facial keypoints.}
\label{tab:abla}
\vspace{-12pt}
\end{table}
\vspace{-1pt}
\subsection{Ablation study}
\begin{figure}[t]
  \centering
   \vspace{-8pt}
  \includegraphics[width=0.95\linewidth]{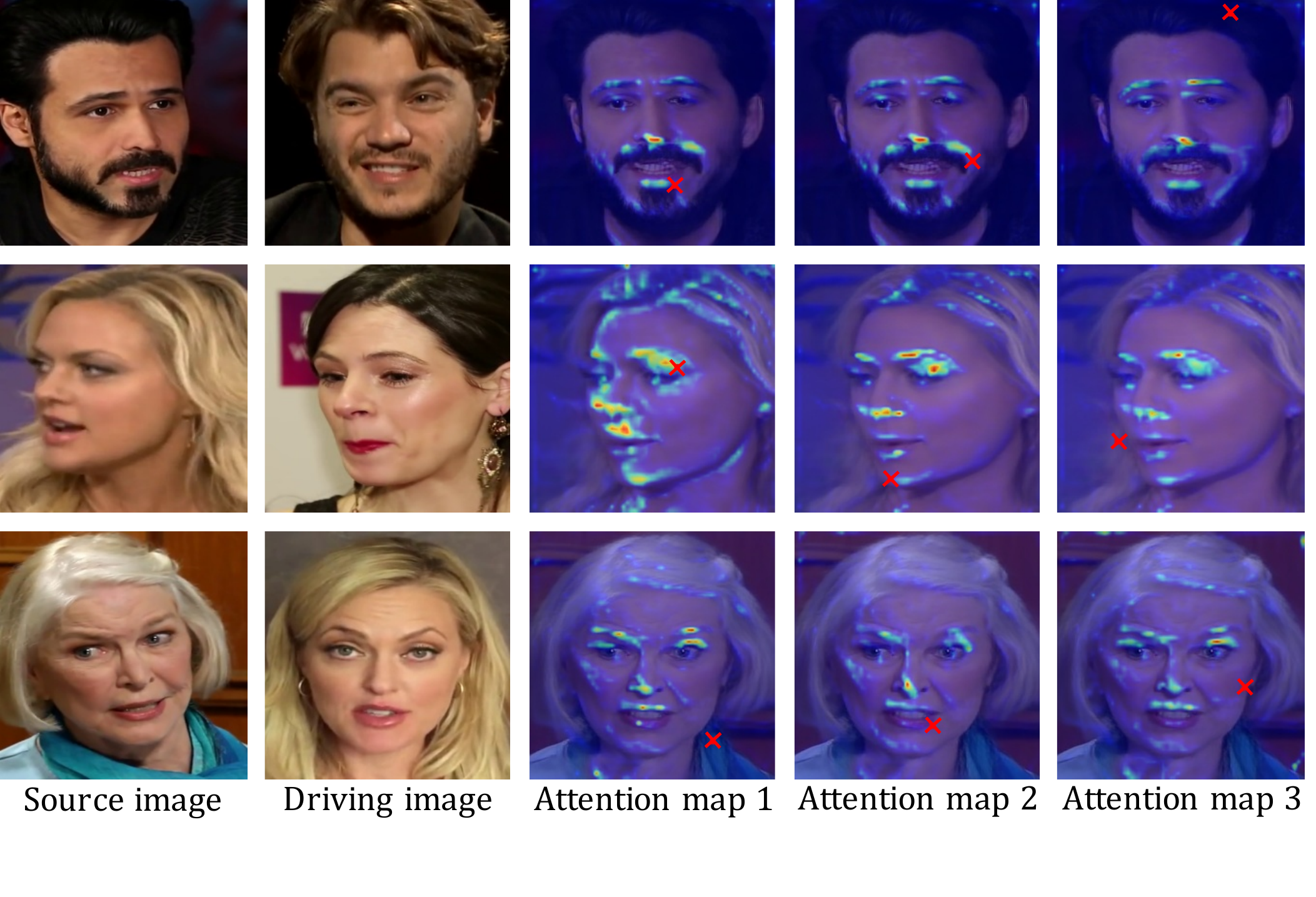}
    \vspace{-20pt}
   \caption{Visualizing the dense depth-aware attention map in cross-modal attention module. In the last three columns, the red mark ``$\times$'' indicates the query location.} 
   \label{fig:attention map}
   \vspace{-13pt}
\end{figure}

\vspace{-0.1cm}
In this section, we conduct ablation studies to demonstrate the effectiveness of the proposed self-supervised face depth learning method and the proposed two mechanisms for talking head generation.
We report results of ablation studies in Tab.~\ref{tab:abla}, and show several qualitative examples of the generation results in Fig.~\ref{fig:abla} and Fig.~\ref{fig:attention map}. 
Here, our baseline is the simplest model trained without the depth map and depth attention module.

\begin{figure}[ht]
    \vspace{-8pt}
    \centering
    \includegraphics[width=0.95\linewidth]{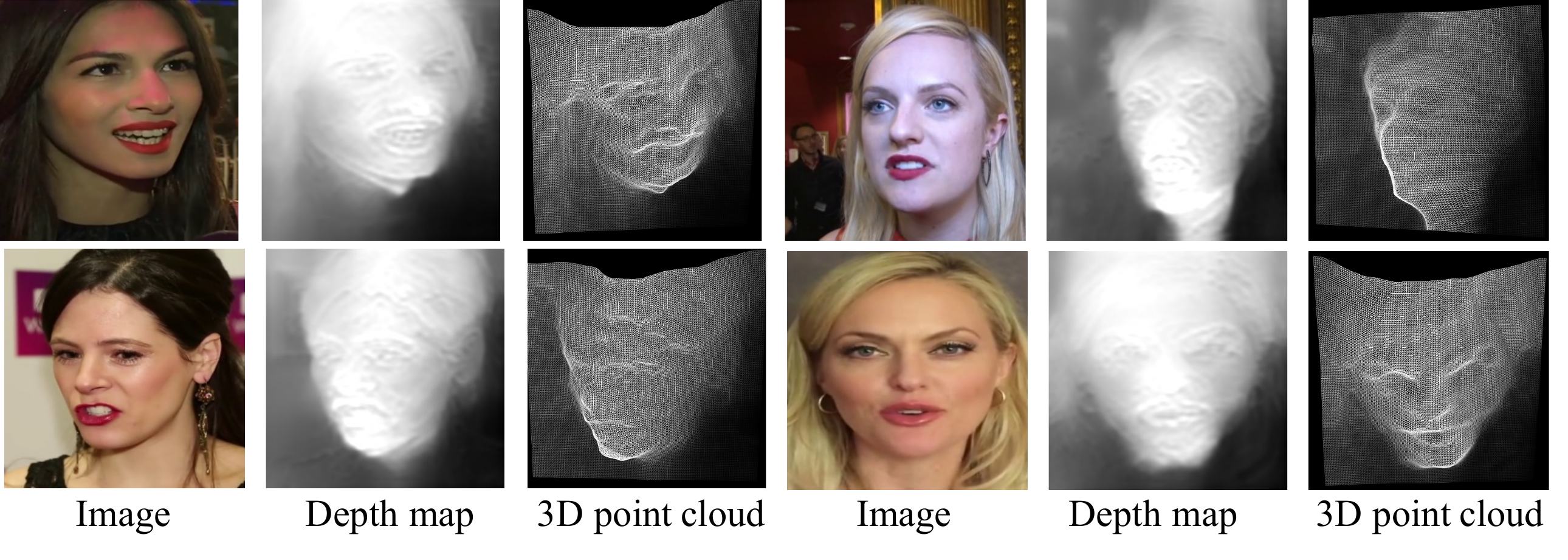}
    \vspace{-10pt}
    \caption{Visualization of estimated face depths and point clouds.}
    \vspace{-0.2cm}
    \label{fig:pointcloud}
\end{figure}
\noindent\textbf{Dense face geometry recovery.} We first show recovered depth maps for human faces from the proposed face depth network. Since we do \emph{not} have any ground-truth depths for the face images, it is tricky to directly evaluate the depth estimation quantitatively. {Thus, we visualize the learned face depth maps and their corresponding 3d point clouds in Fig.~\ref{fig:pointcloud}}. These visualization results strongly demonstrate that our proposed depth learning network is able to effectively recover the dense 3D geometry of human faces.
{The learned dense 3D facial structures are clearly very beneficial, and directly embedded in the proposed model to learn both sparse facial keypoints and global pixel-wise dense attention for the warping of features for generation, leading to a significant improvement on the generation.}

\noindent\textbf{Effectiveness of depth-guided keypionts.}
We aim to explore the impact of depth map on keypoints detection and report the related results in Tab.~\ref{tab:abla}. From Tab.~\ref{tab:abla}, we can easily recognize that the depth-guided keypoints helps our model gain significant gain in PRMSE and AUCON, which indicate that the depth map really plays a significant role in the talking head generation task.~From Fig.~\ref{fig:abla}, the ``Baseline w/ FDN'' predicts more accurate head orientation than ``Baseline'', which can also be observed in Tab.~\ref{tab:abla}, \ie~2.69~vs.~5.39, on the PRMSE metric. {This indicates that the proposed depth-guided facial-keypoints estimator models more accurate motions of the human heads.}
 
\noindent\textbf{Effectiveness of cross-modal attention module.}
From the Tab.~\ref{tab:abla} and Fig.~\ref{fig:abla}, the cross-modal attention module (CAM) can clearly improve the generation quality of expression-related micro-movements of human faces. In Fig.~\ref{fig:abla}, we can observe that the generated face results with the proposed CAM module (\ie~``Baseline w/ CAM'') have more vivid expression (\eg~at eye regions) than that of ``Baseline w/ FDN'' and ``Baseline''. It verifies that the proposed CAM enables the model to capture the expression-related micro-movements at important facial regions (\eg, eyes and mouth). Additionally, the variance ``Baseline w/ CAM'' outperform ``Baseline'' by 0.181 in AUCON.
The results in Tab.~\ref{tab:abla} and Fig.~\ref{fig:abla} verify that our proposed depth attention module can effectively utilize the depth map to enable model focus on micro-movement of the human face to boost the quality of the generated image.

Additionally, we visualize the dense depth-aware attention maps in Fig.~\ref{fig:attention map}. The high activation areas of each query point are mainly located in the expression-related parts of the human face, (\eg eyes, nose, and mouth). These visualization results indicate that our designed cross-modal (\ie~depth and RGB) attention module can indeed address the micro-movements of the human face to produce more vivid expression in generation.


\noindent\textbf{Plug-and-play experiments.} Additionally, we also plug our proposed face depth network and depth-aware cross-modal attention module into FOMM~\cite{siarohin2019first},~\ie, using FOMM as a strong baseline, as our proposed modules can be flexibly deployed into existing video generation methods. The results are reported in Tab.~\ref{tab:abla}. It is obvious that FOMM with the proposed modules can further achieve a significant improvement. These results fully demonstrate the effectiveness of learning dense 3D facial geometry (\ie~depth) for the talking head video generation task.



\vspace{-0.1cm}
\section{Conclusions}
\vspace{-0.1cm}
In this work, we proposed a depth-aware generative adversarial network (DaGAN) for talking head generation. DaGAN learns pixel-wise face depth maps in a self-supervised manner to recover dense 3D facial geometry. We also design two mechanisms to better leverage the depth for the generation. First, we combine the geometry from depth maps and appearance from RGB images to predict more accurate facial keypoints. Second, we design a cross-modal (\ie~depth and RGB) attention mechanism to capture the expression-related micro movements to produce more fine-grained details of facial structures. Ablation studies clearly show that depth maps can benefit the motion transfer between two faces. Our DaGAN also produces more realistic and natural-looking results compared to state-of-the-arts.


{\small
\bibliographystyle{ieee_fullname}
\bibliography{egbib}
}
\newpage
\appendix
\onecolumn
\section*{\LARGE{Appendix}}\label{s:appendix}

\section{Additional Network and Training Details}

\subsection{Loss details}

\noindent\textbf{Perceptual loss $\mathcal{L}_P$.} To ensure that the generated images are similar to their corresponding ground truths, we use a multi-scale implementation introduced by FOMM~\cite{siarohin2019first}. Specifically, we first downsample the ground truth and the output image to 4 different resolutions (\ie $256\times 256$, $128\times 128$, $64\times 64$ and $32\times 32$). We denote $R_1$,$R_2$,$R_3$,$R_4$ as the generated images, and $G_1$,$G_2$,$G_3$,$G_4$ as the corresponding ground truths of the four different resolutions, respectively. Then a pre-trained VGG network is used to extract features from both these downsampled ground truths and the output images. We compute the $\mathcal{L}_1$ distance between the ground truth and output image in different resolutions: 
\begin{equation}
    \mathcal{L}_P = \sum_{i=1}^4 \mathcal{L}_1(G_i,R_i)
\end{equation}

\noindent\textbf{GAN loss $\mathcal{L}_G$.} Given the ground truths and the generated images in $256 \times 256$ resolution, we adopt an adversarial learning objective function consisting of a least square loss and a feature matching loss introduced in the pix2pixHD~\cite{wang2018high} to train our DaGAN. Single-scale discriminators are used for training $256 \times 256$ images.

\noindent\textbf{Equivariance loss $\mathcal{L}_E$.} This loss is utilized to ensure the consistency of the estimated keypoints, which is also adopted by FOMM~\cite{siarohin2019first}. Given an image $\mathbf{I}$ and one of its detected keypoint $\mathbf{x}_k$, we perform a known spatial transformation $\mathbf{T}$ on image $\mathbf{I}$, resulting in a transformed image $\mathbf{I_T}$. Therefore, the detected keypoints $\mathbf{x}_{\mathbf{T}(k)}$ on this transformed image $\mathbf{I_T}$ should be transformed in the same way. Thus, for the $K$ detected keypoints from image $\mathbf{I}$, we have:
\begin{equation}
    \mathcal{L}_E =  \sum_{i=1}^K||\mathbf{x}_k - \mathbf{T}^{-1}(\mathbf{x}_{\mathbf{T}(k)})||_1
\end{equation}

\noindent\textbf{Keypoints distance loss $\mathcal{L}_D$.} To make the detected facial keypoints much less crowded around a small neighbourhood, we employ a keypoints distance loss to penalize the model if the distance between two corresponding keypoints falls below a pre-defined threshold. For every two keypoints $\mathbf{x}_i$ and $\mathbf{x}_j$ in an image, we thus have:
\begin{equation}
    \mathcal{L}_D =  \sum_{i=1}^K\sum_{j=1}^K (1-\mathbf{sign}(||\mathbf{x}_i - \mathbf{x}_j||_1-\alpha)), i\neq j,
\end{equation}
where $\mathbf{sign}(\cdot)$ is a sign function, and the $\alpha$ is the threshold of distance. It is set to 0.2 in our work, which shows good performance in our practice.

\begin{figure*}[t]
  \centering
  \includegraphics[width=1\linewidth]{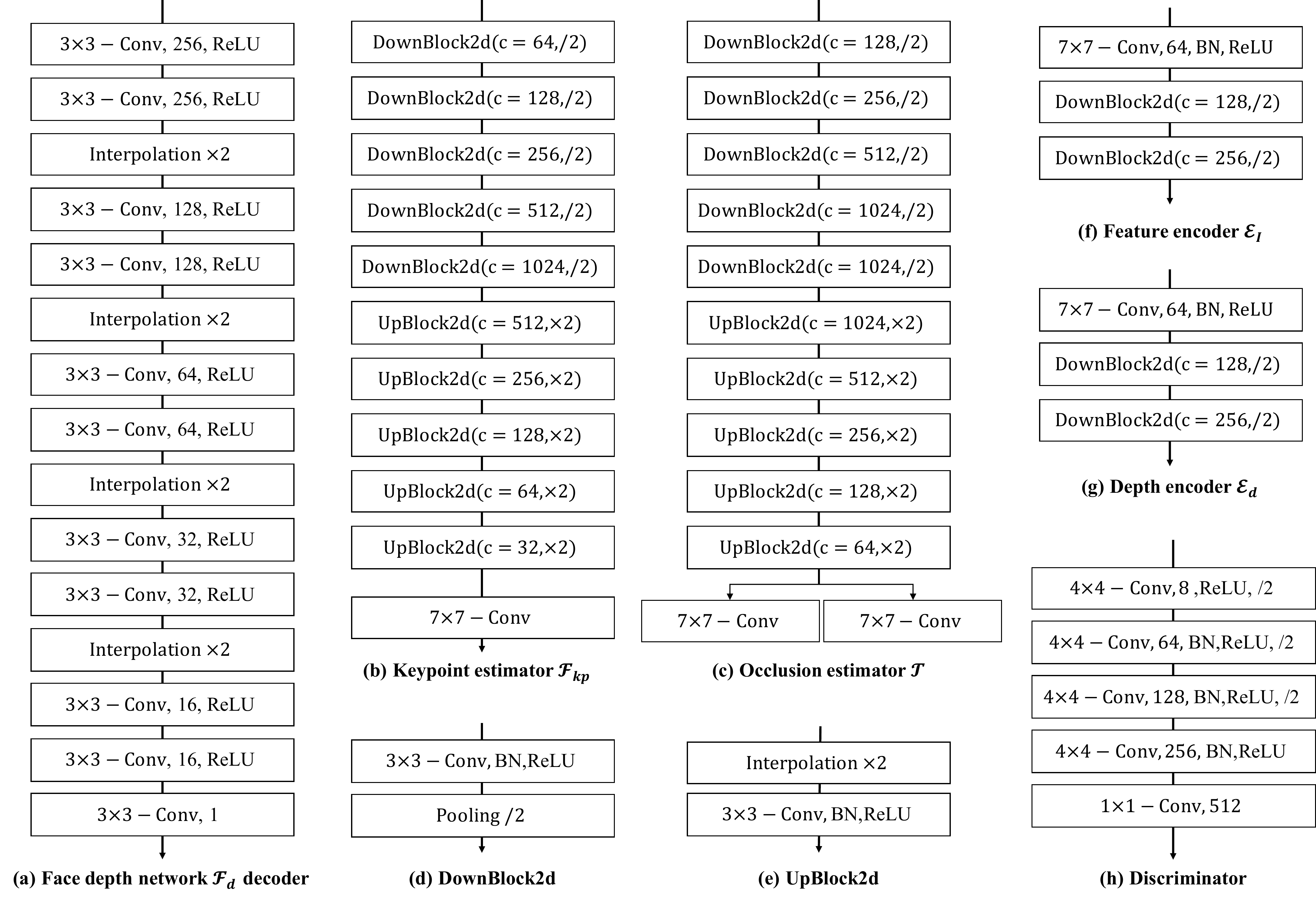}
   \caption{Architecture details of each components in our model. The ``DownBlock2d'' (\cref{fig:structure}d) contains a convolutional layer with $3\times 3$ kernel, a batch normalization layer, a ReLU activation layer, and an average pooling layer that downsamples the input. The interpolation layer in ``UpBlock2d'' (\cref{fig:structure}d) is utilized to upsample the image. The symbol ``/2'' in other 
  sub-networks indicates an average pooling layer to downsample the input.}
   \label{fig:structure}
\end{figure*}
\subsection{Network architecture details of DaGAN}
The implementation details of the sub-networks in our model are shown in~\cref{fig:structure} and described below.

\noindent\textbf{Face depth network $\mathcal{F}_d$.} Our face depth network consists of an encoder and a decoder. The encoder is a ResNet18 network~\cite{he2016deep} without the final fully connected and pooling layers. The structure of the decoder is illustrated in~\cref{fig:structure}a, which predicts a depth map with a size of $1 \times 256 \times 256$.

\noindent\textbf{Keypoint estimator $\mathcal{F}_{kp}$.} In the training process, we concatenate the RGB image and its corresponding depth map to form an RGB-D input with a size of $4 \times 256 \times 256$, while the ouputs are $K$ keypoints $\{\mathbf{x}_{\tau,n}\}_{n=1}^K,\mathbf{x}_{\tau,n}\in \mathbb{R}^{1\times2}$. The detailed structure of the keypoint estimator is shown in~\cref{fig:structure}b.

\noindent\textbf{Occlusion estimator $\mathcal{T}$.} We utilize the occlusion estimator to predict an occlusion map to filter out the regions that should be inpainted, and a motion flow mask for weighting the motion field. As illustrated in~\cref{fig:structure}c, there are two heads at the end to predict these two parts.

\noindent\textbf{Feature encoder $\mathcal{E}_I$.} In~\cref{fig:structure}f, to preserve low-level texture of the image, we only apply two DownBlocks to construct the feature encoder $\mathcal{E}_I$ in the feature warping module. 

\noindent\textbf{Depth encoder $\mathcal{E}_d$.} The architecture of our depth encoder $\mathcal{E}_d$ in the cross-modal attention module is shown in~\cref{fig:structure}g. The structure is the same as $\mathcal{E}_I$, and thus we can make the features learned from both modalities with the same level of representation power. 

\noindent\textbf{Discriminator $\mathcal{D}$.} The architecture of our discriminator (\cref{fig:structure}h) is inspired by FOMM~\cite{siarohin2019first}. The input image is first down-sampled four times, and then passed through a  convolutional layer with a kernel size of $1\times 1$, and we finally output a prediction map with a size of $512 \times 26\times 26$. Moreover, we collect the intermediate feature maps and feed them into the GAN loss $\mathcal{L}_G$. 

\section{Additional Experiment Details}

\subsection{Dataset Details}
\begin{itemize}[leftmargin=*]
\itemsep0em
    \item \textbf{VoxCeleb1} dataset contains videos of 1,251 different identities with a resolution of $256 \times 256$. We extract frames for each video and utilized the test split of VoxCeleb1 for evaluating self-reenactment. Following~\cite{yao2020mesh, ha2020marionette}, we created the test set by sampling 2,083 image sets from randomly selected 100 videos of the VoxCeleb1 test split.
    \item \textbf{CelebV} dataset contains videos of five different celebrities with widely varying characteristics, which are utilize to evaluate the performance of the models for reenacting unseen targets, similar to the in-the-wild scenarios. Moreover, we uniformly sampled 2000 image sets from CelebV to perform the experiments.
\end{itemize}

\subsection{Compare methods}

\begin{itemize}[leftmargin=*]
\itemsep0em
    \item \textbf{X2Face} \cite{wiles2018x2face}. X2Face utilizes a simple framework to warp the image directly. We obtain its results on VoxCeleb1 from a previous work~\cite{ha2020marionette}.
    \item \textbf{NeuralHead}~\cite{zakharov2019few}. NeuralHead adopts an important component from style transfer~\cite{karras2019style, huang2017arbitrary}, \ie AdaIN layers \cite{huang2017arbitrary}. Since a reference implementation is absent, we directly report the replicated results from~\cite{ha2020marionette}.
    \item \textbf{MarioNETte} \cite{ha2020marionette}. MarioNETte utilizes three components (\ie image attention block, target feature alignment, and landmark transformer) to address the identity preservation problem. We compare with it based on the results reported in the original paper.
    \item \textbf{FOMM}~\cite{siarohin2019first}. FOMM propose a paradigm that aims to detect the keypoints of the face image and model the motion between two images using detected keypoints.
    \item \textbf{MeshG}~\cite{yao2020mesh}. MeshG aims to generate a dense face mesh to model a dense motion map using graph convolutional network. As there is no official code available, we only report the its results from the original paper.
    \item \textbf{OSFV}~\cite{wang2021one}. OSFV provides a novel keypoint generation method. We reimplemented this method according to its published paper and train it on the VoxCeleb1 dataset to compare with the proposed method.
\end{itemize}


\subsection{More results}
\begin{figure}[h]
    \vspace{-10pt}
    \centering
    \includegraphics[width=0.97\linewidth]{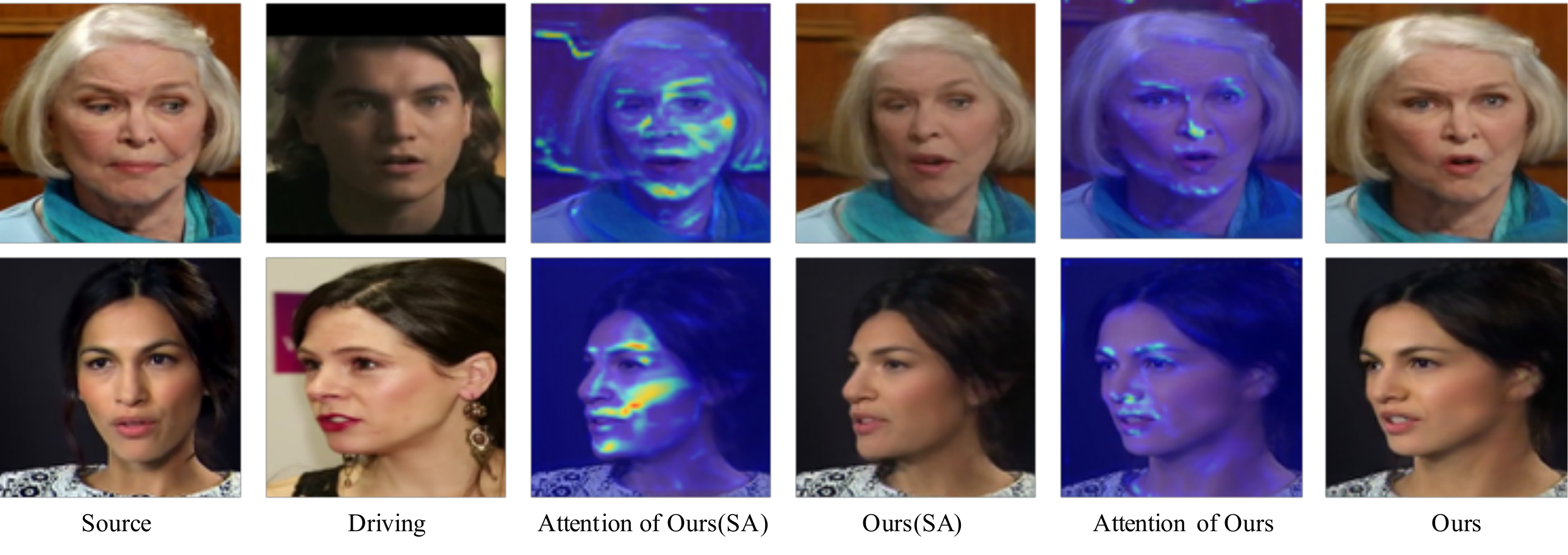}
    \vspace{-12pt}
    \caption{Visualization of attention maps of different methods.}
    \vspace{-0.4cm}
    \label{fig:attn_compare}
\end{figure}
\noindent\textbf{More explanation of depth-aware attention.}
Each learned 3D spatial depth point is inherently used as a query for calculating a global self-attention, which is thus depth-aware. Here, we disable the depth in the cross-modal attention module, which then becomes a standard self-attention module, termed as Ours~(SA).
A qualitative comparison in Fig.~\ref{fig:attn_compare} shows the difference of using and not using depth for the attention learning. Our cross-modal attention can effectively learn to attend to key foreground facial regions (e.g.~expression-related keypoint regions), comparing to the one without depth (\ie~Ours~(SA)) which also attends to cluttered backgrounds, further confirming the advantage of dense 3D geometry for overcoming noisy background in generation.

\begin{figure*}[h]
  \centering
  \includegraphics[width=1\linewidth]{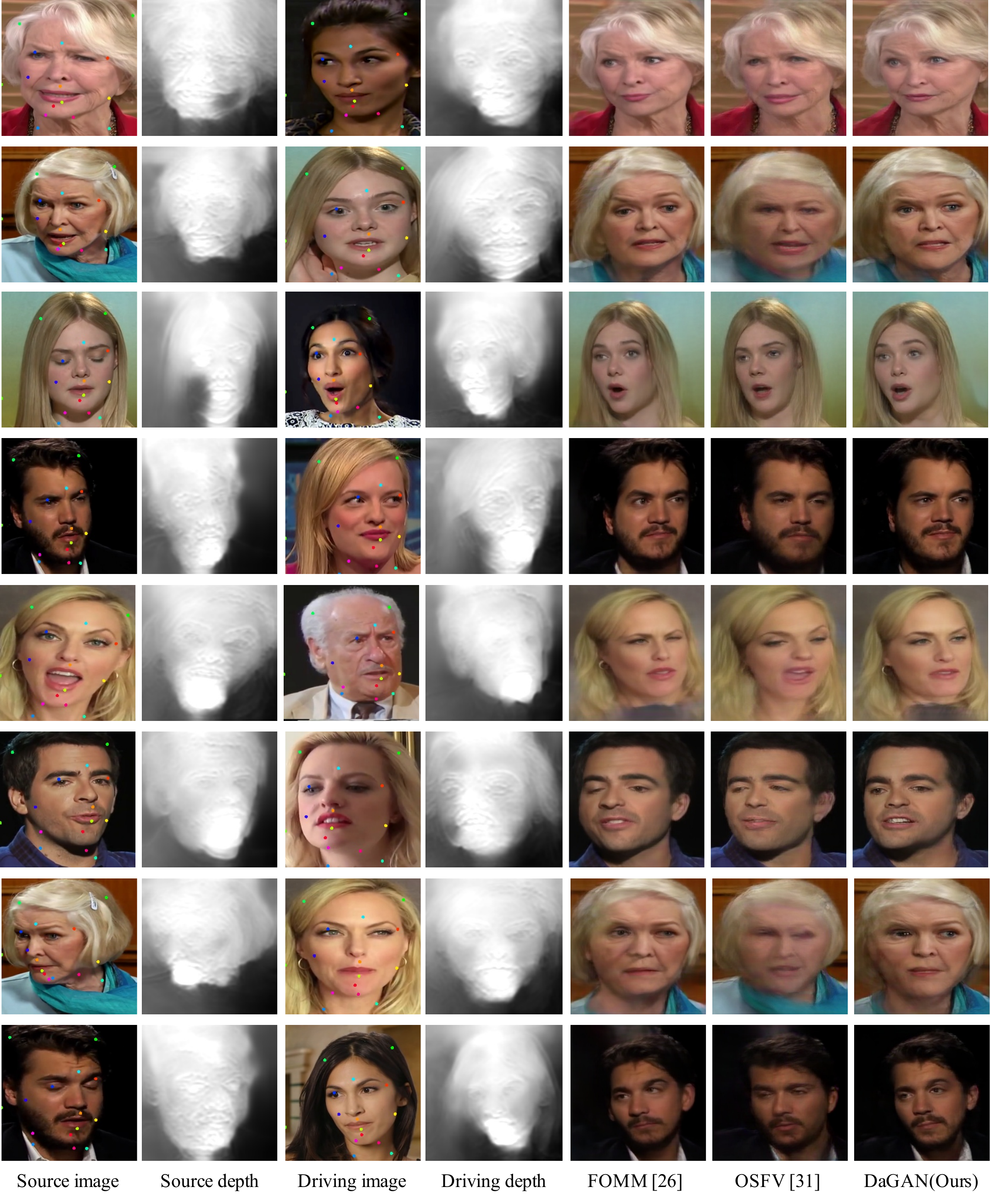}
  \vspace{-20pt}
   \caption{Qualitative comparisons of different methods on cross-identity face reenactment. We also show the predicted face depth maps and detected keypoints of source images and driving images.}
   \label{fig:supp_vox1}
\end{figure*}

\begin{figure*}[h]
  \centering
  \includegraphics[width=1\linewidth]{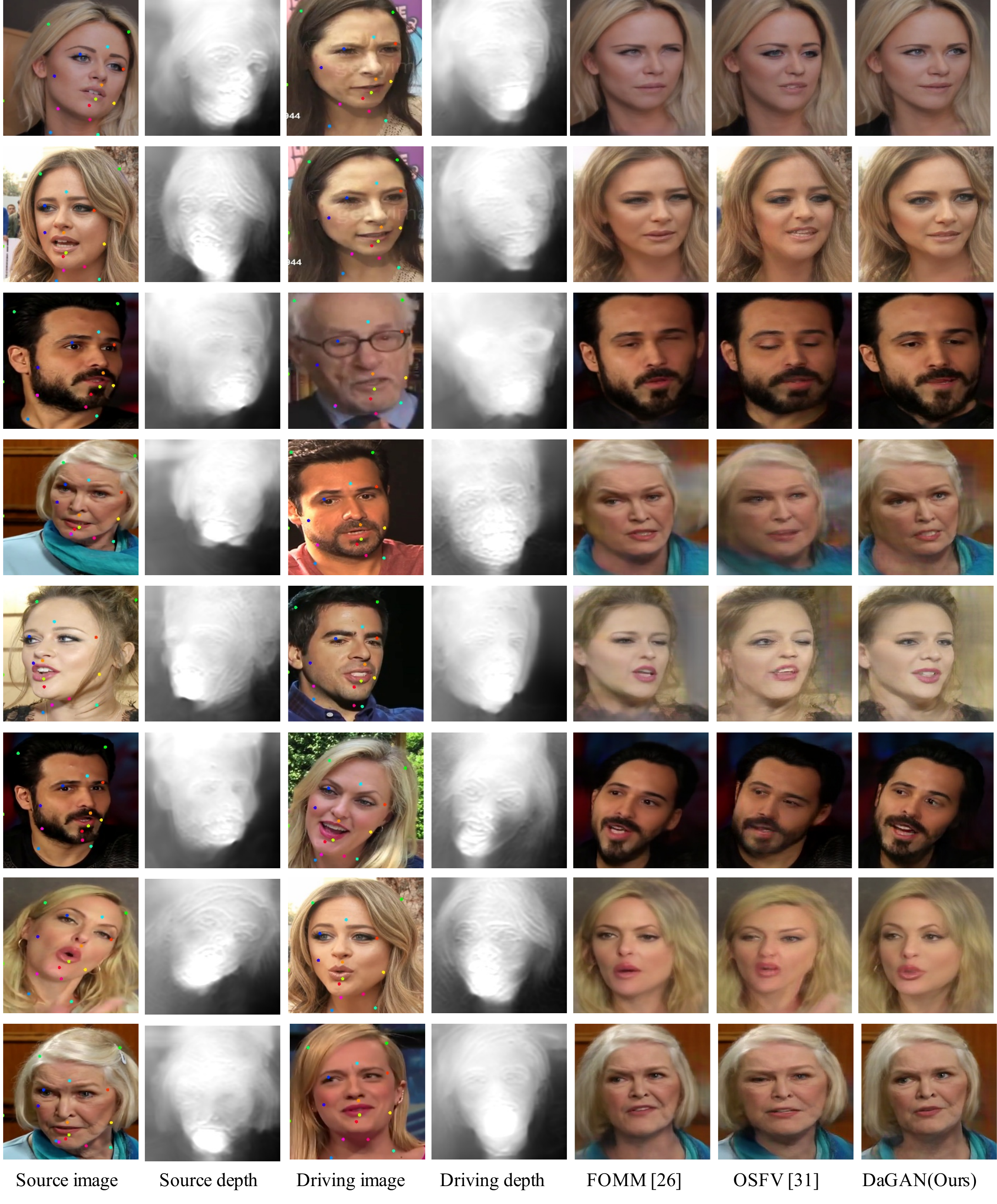}
  \vspace{-20pt}
   \caption{Qualitative comparisons of different methods on cross-identity face reenactment. We also show the predicted face depth maps and detected keypoints of source images and driving images.}
   \label{fig:supp_vox2}
\end{figure*}

\noindent\textbf{More qualitative results.}
We show more samples in \cref{fig:supp_vox1} and \cref{fig:supp_vox2}. The visualization shows that our DaGAN can produce more natural-looking faces than the other comparison methods. More than that, we also present our generated depth maps of the source images and the driving images. We can observe that our estimated depth maps can effectively distinguish the face foreground area of an image from the background. 
These robustly predicted depth maps can also verify the effectiveness of our method for self-supervised dense geometry recovery.
\par\noindent\textbf{Video generation demo.} We also provide a video generation demonstration to show a more detailed comparison qualitatively with the most competitive methods in the literature, including FOMM~\cite{siarohin2019first} and OSFV~\cite{wang2021one}. The demo is attached together with this supplement document.

\end{document}